\RequirePackage{fix-cm}
\RequirePackage{amsmath}

\documentclass[
onecolumn
]{svjour3}          % twocolumn

\smartqed  % flush right qed marks, e.g. at end of proof
\usepackage{graphicx}
\usepackage[utf8]{inputenc}
\usepackage[T1]{fontenc}
\usepackage{hyperref}
\usepackage{booktabs}
\usepackage[percent]{overpic}

\usepackage{tabularx}
\newcolumntype{C}{>{\centering\arraybackslash}X}

\usepackage{multirow}
\usepackage{amssymb}
\usepackage{textcomp}

\usepackage[group-separator={,}]{siunitx}
\usepackage{tikz}
\journalname{IJCARS}
\begin{document}

\title{Endoscopic vs.\ volumetric OCT imaging of mastoid bone structure for pose estimation in minimally invasive cochlear implant surgery}

\titlerunning{Endoscopic vs.\ OCT imaging of mastoid structure for pose estimation in MICI}

\author{Max-Heinrich Laves \and
        Sarah Latus \and
        Jan Bergmeier \and
        Tobias Ortmaier \and
        Lüder A. Kahrs \and
        Alexander Schlaefer
}


\institute{M.-H.\ Laves \and J. Bergmeier \and T. Ortmaier \and L. A. Kahrs\at
              Institute of Mechatronic Systems,\\ Leibniz Universit\"at Hannover, Germany\\
              Tel.: +49-511-762-19617\\
              Fax: +49-511-762-19976\\
              \email{laves@imes.uni-hannover.de} \\
            \and
            S. Latus \and A. Schlaefer \at
            Institute of Medical Technology,\\ Hamburg University of Technology, Hamburg, Germany
}

%\institute{Max-Heinrich Laves
%\and Lüder A. Kahrs \and Tobias Ortmaier \at
%           Appelstra\ss{}e 11A \\
%           30167 Hannover, Germany \\
%           Tel.: +49\,511 762\,19617 \\
%           Fax: +49\,511 762\,19976 \\
%           \email{laves@imes.uni-hannover.de}           %  \\
%           \emph{Present address:} of F. Author  %  if needed
%}

\date{Received: date / Accepted: date}

\maketitle

\begin{abstract}
\textit{Purpose} The facial recess is a delicate structure that must be protected in minimally invasive cochlear implant surgery.
Current research estimates the drill trajectory by using endoscopy of the unique mastoid patterns.
However, missing depth information limits available features for a registration to preoperative CT data.
Therefore, this paper evaluates OCT for enhanced imaging of drill holes in mastoid bone and compares OCT data to original endoscopic images. \\
\textit{Methods} A catheter-based OCT probe is inserted into a drill trajectory of a mastoid phantom in a translation-rotation manner to acquire the inner surface state.
The images are undistorted and stitched to create volumentric data of the drill hole.
The mastoid cell pattern is segmented automatically and compared to ground truth. \\
\textit{Results} The mastoid pattern segmented on images acquired with OCT show a similarity of $ J = 73.6\,\% $ to ground truth based on endoscopic images and measured with the Jaccard metric.
Leveraged by additional depth information, automated segmentation tends to be more robust and fail-safe compared to endoscopic images. \\
\textit{Conclusion} The feasibility of using a clinically approved OCT probe for imaging the drill hole in cochlear implantation is shown.
The resulting volumentric images provide additional information on the shape of caveties in the bone structure, which will be useful for image-to-patient registration and to estimate the drill trajectory.
This will be another step towards safe minimally invasive cochlear implantation.

\end{abstract}

\section{Introduction}

Safe drilling in minimally invasive cochlear implantation (MICI) needs verification of the drill trajectory before passing the facial recess by intraoperative imaging or other sensor data \cite{Labadie2013,Weber2017}.
The X-ray-based and costly gold standard method using an intraoperative computed tomography (CT) scan can be replaced by using optical modalities, like endoscopy.
This provides visual access to the inner drill hole surface, but the stitched two-dimensional endoscopy images provide no corresponding volumetric information of surfaces visible from within the drill hole.
Nevertheless, image-to-patient registration is feasible with the endoscopic and preoperative CT data \cite{Bergmeier2017}.
The features used for this registration and pose estimation of the trajectory derive from the pattern of bone and air-pockets in the mastoid, intersected by the drill trajectory.
Volumetric information from the optical imaging modality is anticipated to raise the accuracy of the registration method.

In previous publications, we have demonstrated that optical coherence tomography (OCT) is useful for intraoperative guidance in head surgery \cite{Diaz-Diaz2013} as well as for pose estimation \cite{Gessert2018}.
The potential of using OCT for precise navigation in bony structures has also been illustrated by Zhang et al. \cite{Zhang2014}.
Additionally, we investigated catheter-guided OCT imaging and related volume reconstructions methods \cite{Latus2018}.
This paper evaluates to what extend OCT can be used to acquire volumetric images of the inner drill hole in mastoid structure during MICI for future pose estimation of the drill trajectory and image-to-patient registration.
We compare the stitched image information of our endoscopic and the OCT exploration technique. 

\section{Methods}

% \begin{figure}
%   \centering
%   \begin{minipage}[c][1\width]{0.49\columnwidth}%
%   \includegraphics[width=0.95\columnwidth]{Felsenbein_Katheter_small.jpg} ~
%   \end{minipage}
%   \begin{minipage}[c][1\width]{0.49\columnwidth}%
%   \includegraphics[width=0.95\columnwidth]{Katheter_small.jpg}
%   \end{minipage}
%   \caption{Left: Mastoid phantom with drill hole (red arrow). Several holes with different orientations w.r.t. the drill hole are drilled in the phantom to simulate the air-pockets in the mastoid. The imaging catheter is positioned straight and centered to the drill hole using a glass capillary (blue arrow).
%   Right: OCT imaging catheter positioned within the glass capillary. The OCT beam of the imaging probe (red arrow) is oriented perpendicular to the catheter axis. The probe is rotated and pulled back step-wise in order to acquire radial depth profiles of the mastoid phantom.} %TODO SL: FIGURE BESCHRIFTEN
%   \label{fig:experimental_setup}
% \end{figure}

\subsection{Experimental setup}

\begin{figure}[htb]
    \centering
    \includegraphics{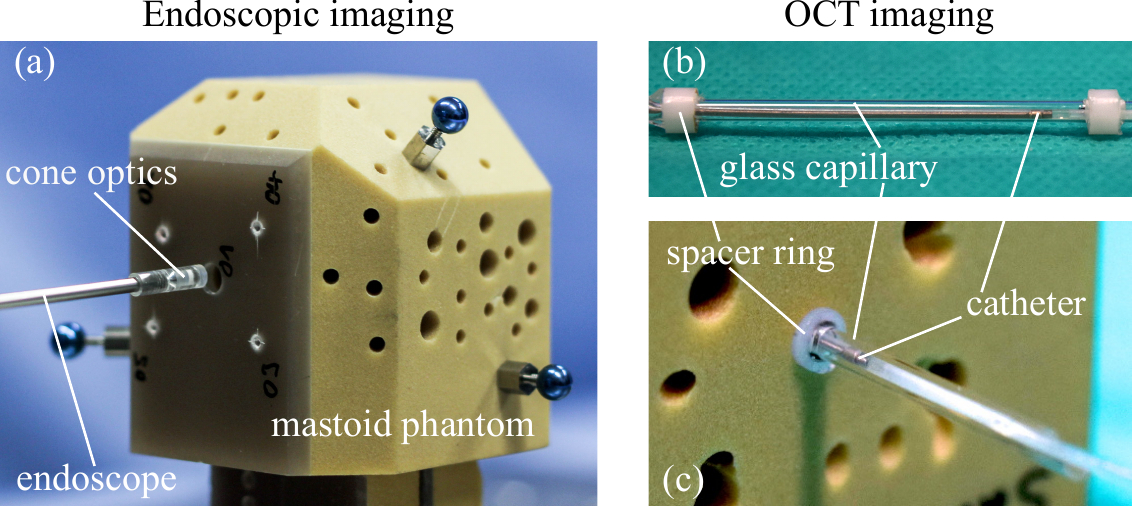}
    \caption{(a) Mastoid phantom with drill hole. Several holes with different diameters and orientations are drilled into the phantom to simulate the air-pockets in the mastoid. The endoscope, extended with the cone
optics is used for drill hole imaging. (b) OCT imaging catheter positioned within the glass capillary. (c) Imaging catheter, positioned concentric in the drill hole, using a glass capillary and spacer rings.}
    \label{fig:experimental_setup}
\end{figure}

A mastoid phantom consisting of bone substitute material was used for the experiments (see Fig.~\ref{fig:experimental_setup}).
For the OCT measurements, a clinically approved imaging catheter (Dragonfly Duo Kit, Abbott) with an outer diameter of 0.9\,mm is connected via a custom made adapter to a spectral-domain OCT device (Telesto I, Thorlabs) with an A-scan rate of $ f = 91 $\,kHz, as proposed in \cite{Latus2018}.
The catheter is positioned straight in the drill hole using a glass capillary which is fitted in the drill trajectory with two rings at both ends (see Fig.~\ref{fig:experimental_setup}).

%\begin{figure}[tb]
%    \centering
%    \includegraphics[scale=0.1]{oct_setup.JPG}
%    \caption{Caption.}
%    \label{fig:experimental_setup2}
%\end{figure}
% The imaging catheter is positioned concentric in the drill hole using a glass capillary. The OCT beam of the imaging probe is oriented perpendicular to the catheter axis. The probe is rotated and pulled back step-wise in order to acquire radial depth profiles of the mastoid phantom.

The imaging probe within the catheter is rotated by the custom made adapter at 390\,rpm to obtain radial A-scans at a rate of 91\,kHz, resulting in B-scans (OCT slice) comprising 14k A-scans per full rotation. After acquisition of an OCT slice, the imaging probe is pulled backwards by $ d = \SI{200}{\micro\meter} $ to acquire the next slice.
This procedure is repeated over a pullback length of \SI{30}{\milli\meter}, resulting in an image with 150 slices.

The endoscopic image data for comparison is recorded using a straight viewing Hopkins II Telescope (KARL STORZ GmbH \& Co.\ KG, Tuttlingen, Germany), extended by a cone mirror setup, as described before \cite{Bergmeier2017a}. A reflective cone surface provides a \SI{360}{\degree} orthogonal view onto the drill hole surface by reflecting perpendicular incoming light-rays into the endoscope optics. Similar to the OCT-recording, consecutive frames are recorded while moving the endoscope through the drill hole. Each frame is unrolled into an image-stripe, performing a transformation from polar to Cartesian coordinates. Considering the visible area and the feed motion, all stripes are stitched to generate the panoramic view of the drill hole surface.

\subsection{Image undistortion}

Due to the catheter not being perfectly centrically aligned in the capillary glass, the resulting unrolled OCT slices show sinusoid distortions.
In order to correct this, the following imaging pipeline is employed.
In every OCT slice, the glass-air transition of the capillary tube and drill hole cavity is clearly visible and appears to be with sinusoid profile depicting the collinear misalignment (see Fig.~\ref{fig:mscan}).
To segment it, the parameters $ A, \omega, \varphi, D $ of the sine wave function
\begin{equation}
%	f: \mathbb{R} \rightarrow \mathbb{R}, \quad
	f(x) = A\sin(\omega x + \varphi) + D
\end{equation}
are optimized by using a RANSAC robust estimator with manually estimated starting values.
The data used for this optimization are the $ u,v $ coordinates of all non-zero pixels after local threshold binarization.
To make the thresholding more robust, image enhancement is applied by wavelet-based denoising and contrast limited histogram equalization.
The final image is undistorted by the inverse of the found sine wave function.

\subsection{Extracting mastoid pattern}

The mastoid pattern is extracted by cropping the undistorted image and performing a maximum intensity projection (MIP) along the vertical image axis.
Using the maximum value indices, a 1D signal is obtained that represents the depth information of the drill hole surface at one axial position.
This is repeated for each of the slices individually.
Every 1D signal forms a row of a new image, which in the end shows the complete unrolled surface of the drill hole.

Additionally, the undistorted and cropped slices can directly be used to reconstruct a dense 3D hollow cylinder, which also shows subsurface information.
The images can be exploited for 2D-3D or 3D-3D image-to-patient registration with preoperative CT scans, respectively.

\section{Results}
\label{sec:results}

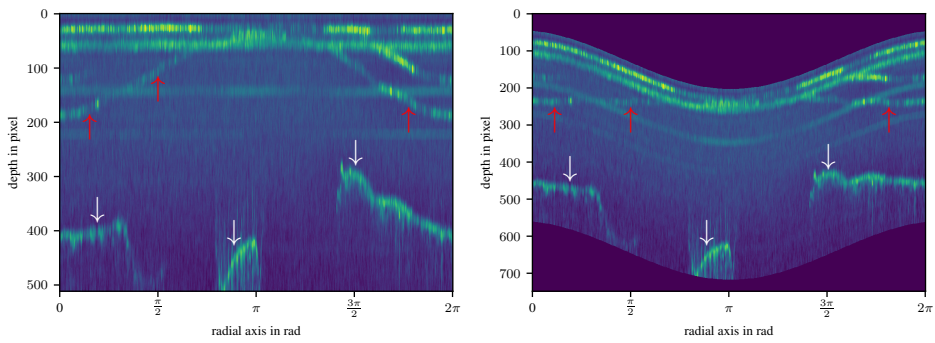
\begin{figure*}[tb]
  \centering
  \begin{tikzpicture}
    \draw (0, 0) node[inner sep=0] {\includegraphics[width=0.49\columnwidth]{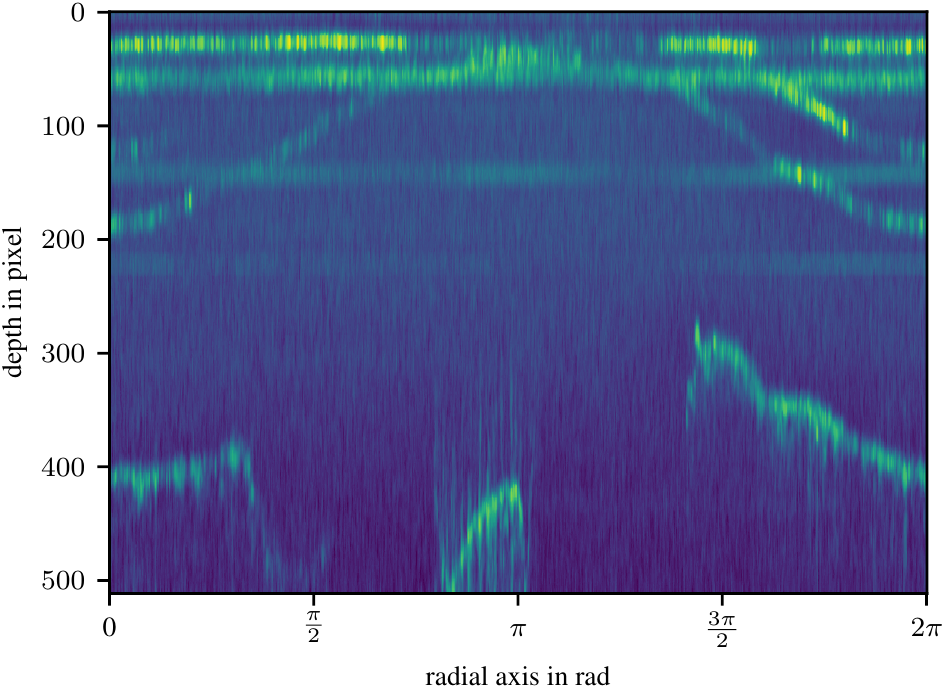}};
    \draw (-1.9, 0.6) node {\color{red}\large $\uparrow$};
    \draw (-1.0, 1.1) node {\color{red}\large $\uparrow$};
    \draw (2.3, 0.7) node {\color{red}\large $\uparrow$};
    
    \draw (-1.8, -0.5) node {\color{white}\large $\downarrow$};
    \draw (-0.0, -0.8) node {\color{white}\large $\downarrow$};
    \draw (1.6, 0.25) node {\color{white}\large $\downarrow$};
  \end{tikzpicture}
  \hfill
  \begin{tikzpicture}
    \draw (0, 0) node[inner sep=0] {\includegraphics[width=0.49\columnwidth]{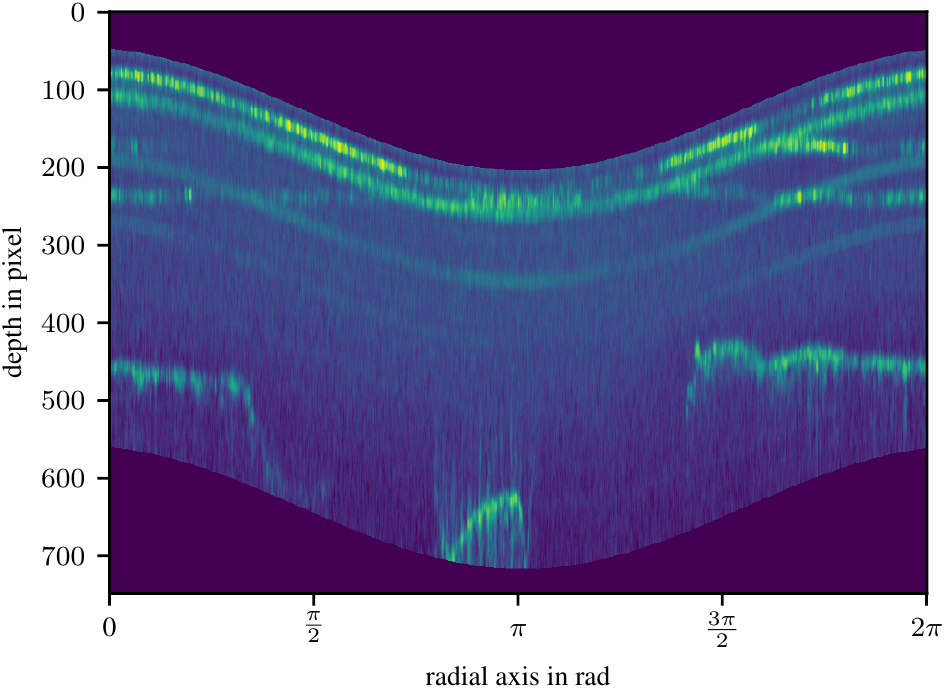}};
    \draw (-2.0, 0.7) node {\color{red}\large $\uparrow$};
    \draw (-1.0, 0.7) node {\color{red}\large $\uparrow$};
    \draw (2.4, 0.7) node {\color{red}\large $\uparrow$};
    
    \draw (-1.8, 0.05) node {\color{white}\large $\downarrow$};
    \draw (-0.0, -0.8) node {\color{white}\large $\downarrow$};
    \draw (1.6, 0.2) node {\color{white}\large $\downarrow$};
  \end{tikzpicture}
  \caption{Left: OCT B-scan in polar coordinates (unrolled). The sinusoid layer (red arrows) represents the outer capillary glass borderline. The drill hole surface is visible at the bottom (white arrows). Right: B-scan undistorted by sinusoid correction. The outer capillary glass borderline appears to be horizontal after undistortion.}
  \label{fig:mscan}
\end{figure*}

\begin{figure*}[tb]
  \centering
  \begin{overpic}[width=0.19\columnwidth]{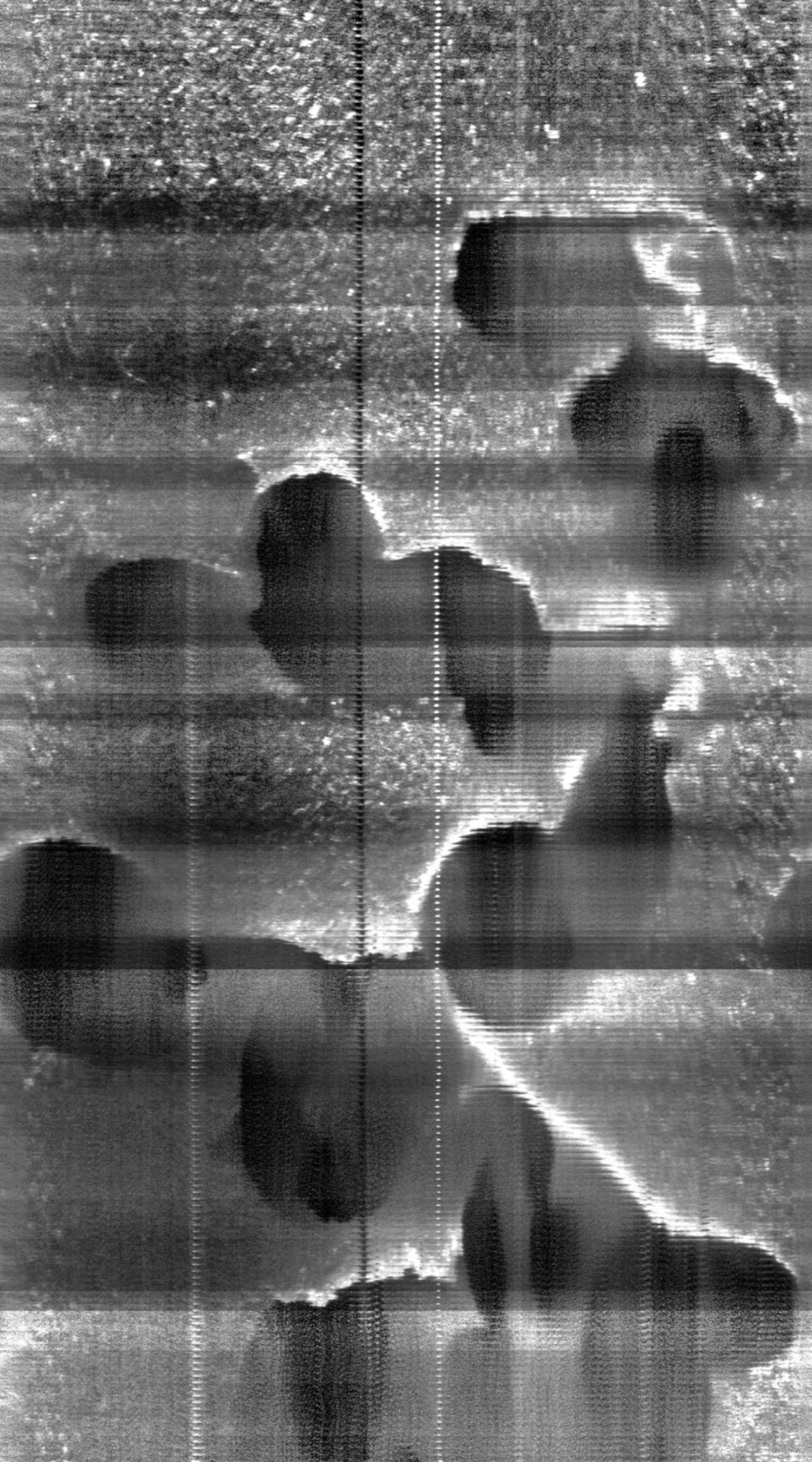}
    \put (1, 3.5) {\setlength{\fboxsep}{1pt}\colorbox{white}{\footnotesize (a)}}
  \end{overpic} \hfill
  \begin{overpic}[width=0.19\columnwidth]{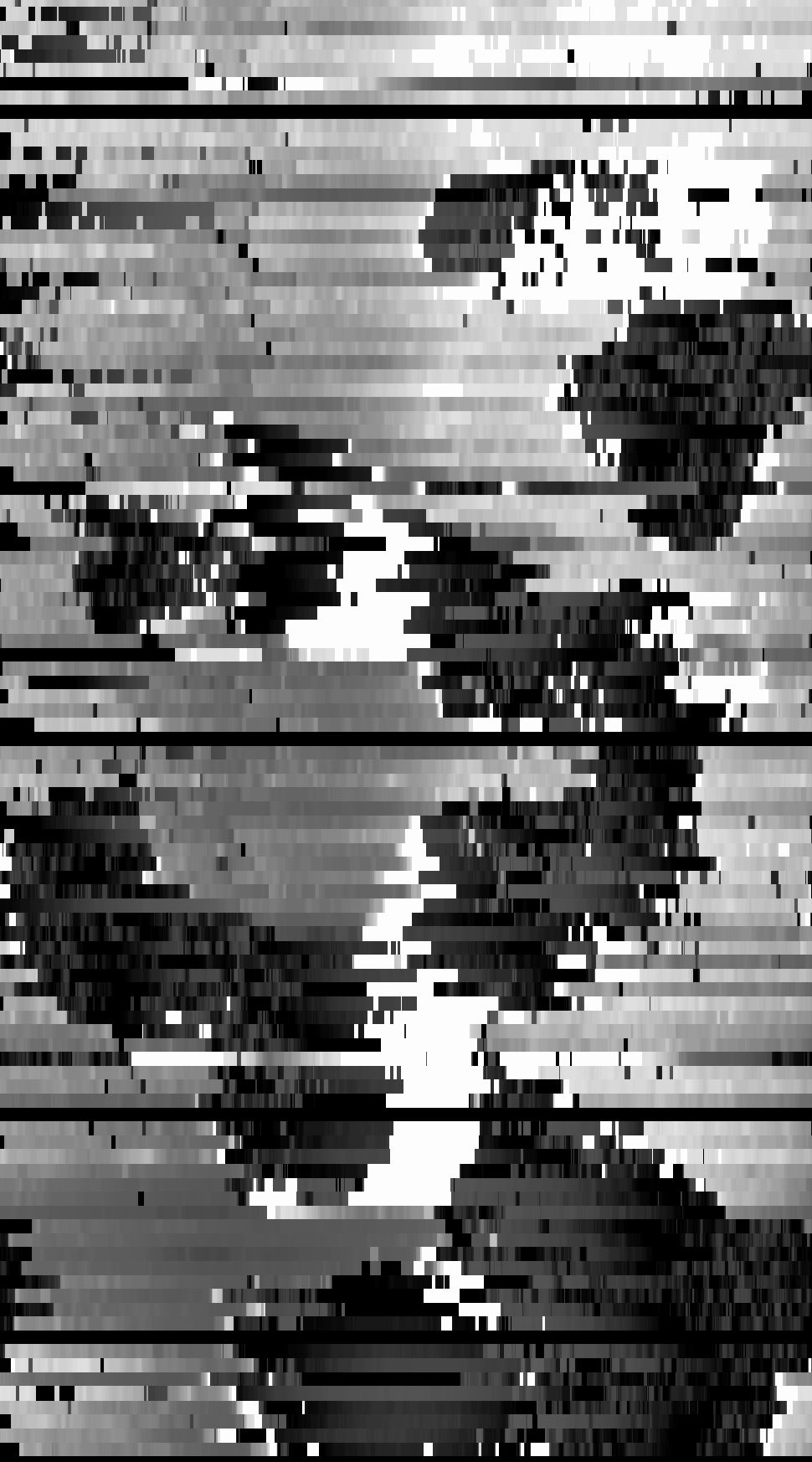}
    \put (1, 3.5) {\setlength{\fboxsep}{1pt}\colorbox{white}{\footnotesize (b)}}
  \end{overpic} \hfill
  \begin{overpic}[width=0.19\columnwidth]{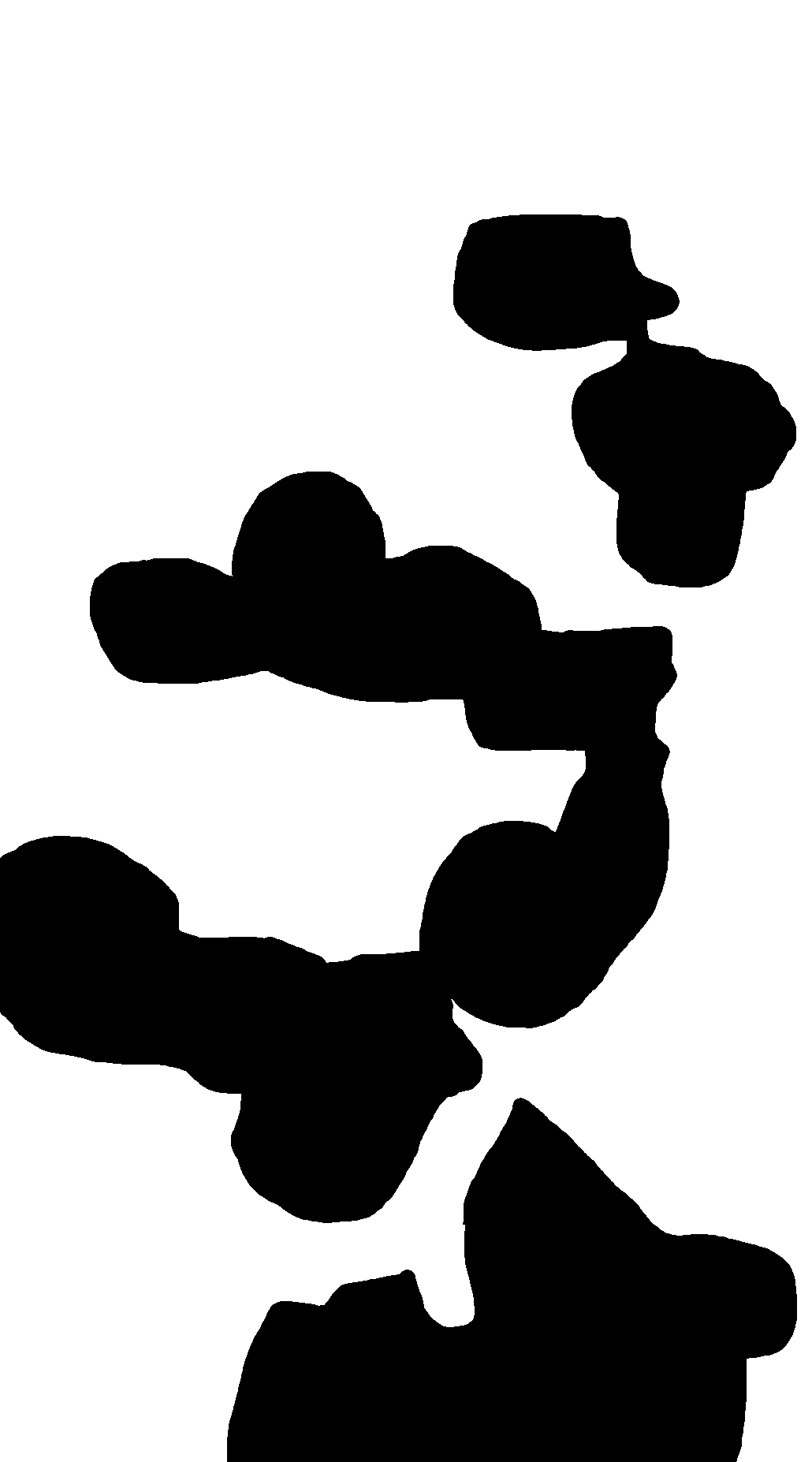}
    \put (1, 3.5) {\setlength{\fboxsep}{1pt}\colorbox{white}{\footnotesize (c)}}
  \end{overpic} \hfill
  \begin{overpic}[width=0.19\columnwidth]{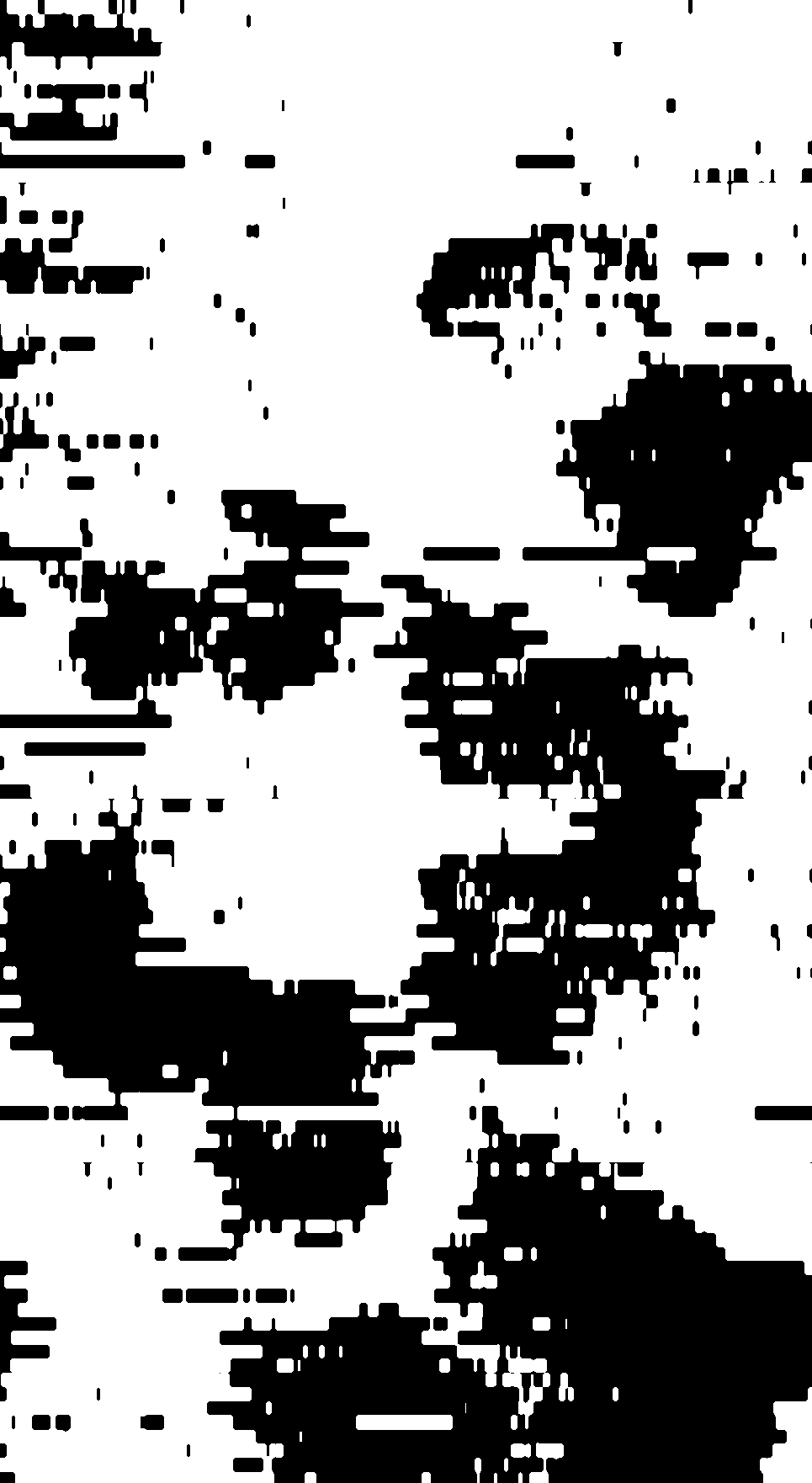}
    \put (1, 3.5) {\setlength{\fboxsep}{1pt}\colorbox{white}{\footnotesize (d)}}
  \end{overpic} \hfill
  \begin{overpic}[width=0.19\columnwidth]{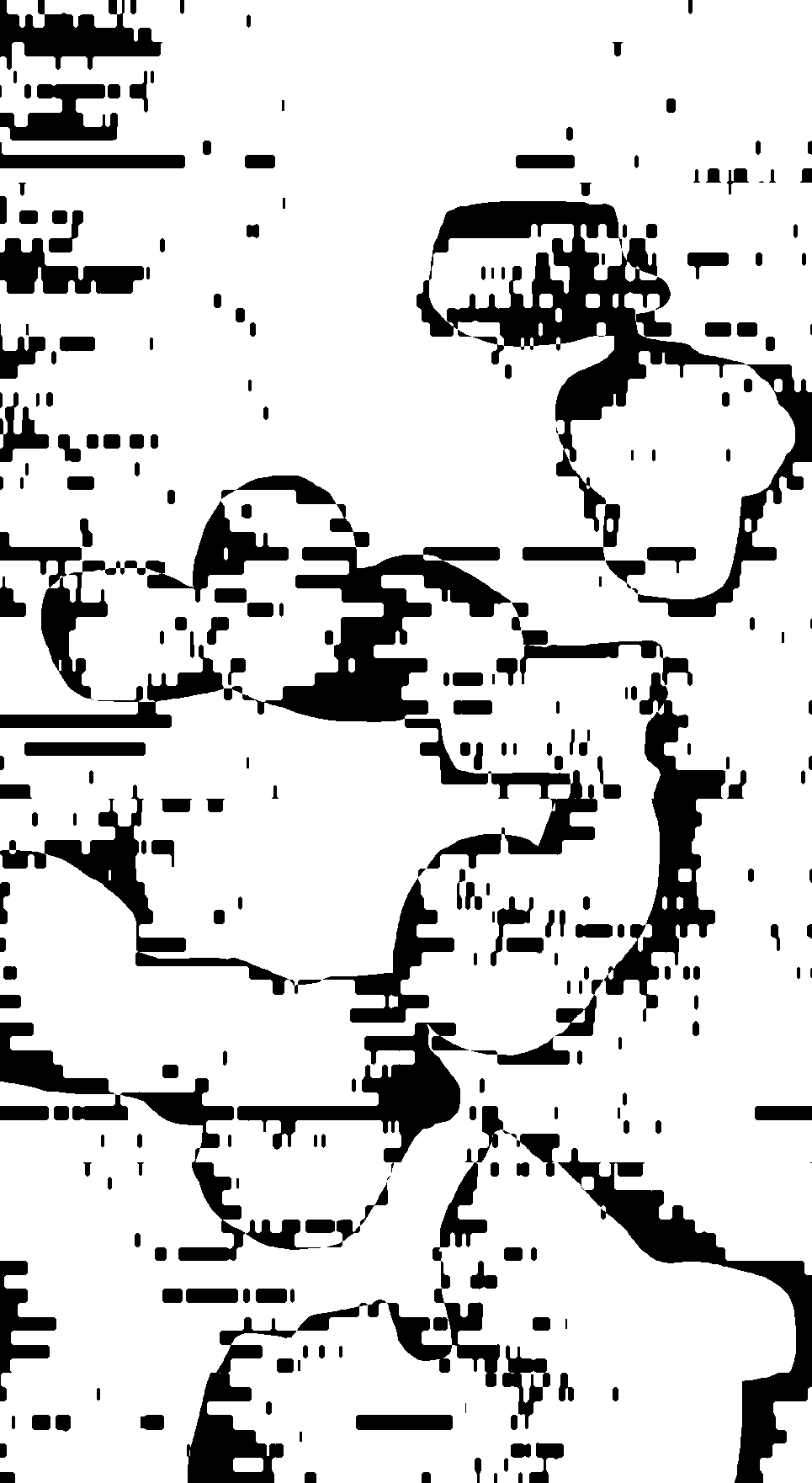}
    \put (1, 3.5) {\setlength{\fboxsep}{1pt}\colorbox{white}{\footnotesize (e)}}
  \end{overpic}
  \caption{Comparison of unrolled endoscopic (a) and OCT (b) image of the drill trajectory. Segmented mastoid patterns from manually delineated endoscopic ground truth (c) and OCT (d) image as well as resulting difference image (e).} %??? endoscopic image as gt??
  \label{fig:unrolled}
\end{figure*}

\begin{figure*}[htb]
    \centering
    \includegraphics[width=0.9\textwidth]{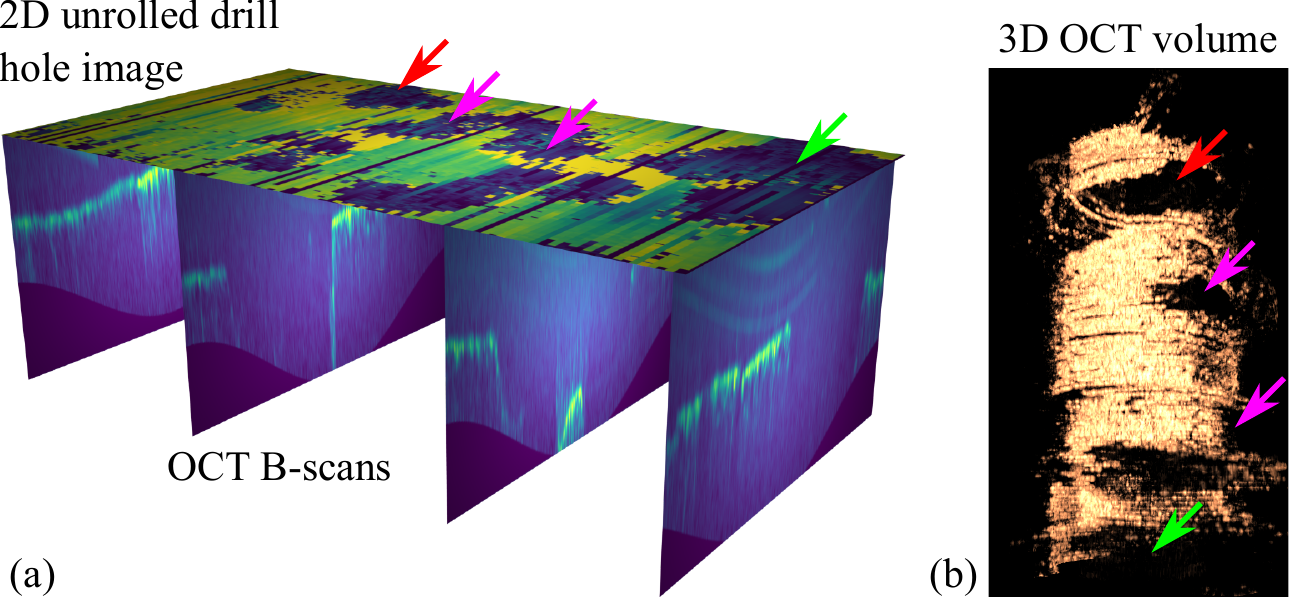}
    \caption{(a) Unrolled drill hole image (as shown in Fig.~\ref{fig:unrolled} (b)) with color-coded depth. OCT B-scans at different axial positions showing the underlying depth profile. B-scan colors correspond to the signal intensity. (b) Cylindrical OCT volume rendering of the drill hole. The colored arrows show corresponding mastoid structures.}
    \label{fig:rendering}
\end{figure*}

A raw B-scan at one axial position and the respective undistorted image are shown in Fig.~\ref{fig:mscan}.
The result of the presented approach is shown in Fig.~\ref{fig:unrolled}~(b), in which the 1D depth information are stacked to compose the final OCT image of the drill hole.
The mastoid patterns are clearly visible and due to the depth information of the OCT B-scans a robust recognition of the air-pockets is possible.
Fig.~\ref{fig:rendering} emphasizes the use of available depth information.

To evaluate the approach quantitatively, the mastoid patterns are segmented employing minimum cross entropy thresholding \cite{Li1998} (see Fig.~\ref{fig:unrolled} (d)).
Next, the Jaccard similarity coefficient
\begin{equation}
    J(A,B) = \frac{\vert A \cap B \vert}{\vert A \cup B \vert}
\end{equation}
is calculated between the binarized mastoid pattern sets $ A $ acquired with the endoscope (Fig.~\ref{fig:unrolled}~(c)) and $ B $ acquired with the catheter-guided OCT (Fig.~\ref{fig:unrolled}~(d)).
The presented approach is able to image the inner drill hole and with $ J = 73.6 \,\% $ and the overlap shows significant similarity (see Fig.~\ref{fig:unrolled} (e)).
Note, that the similarity is computed with respect to the endoscopic images, where depth is not is not considered and may lead to artifacts.

\section{Conclusion}
\label{sec:conclusion}

MICI depends on safety methods to ensure a passage through the facial recess without harming any vital structures. Beyond our existing endoscopic pose estimation and image-to-patient registration technique \cite{Bergmeier2017} this paper compares and extends the input image for this method with a new approach based on catheter-guided OCT. Feasibility of OCT imaging in the drill hole for MICI was proven with experiments based on a mastoid bone phantom. High similarities of catheter-based OCT and endoscopic images were shown. 

Next steps of our research will include using the OCT image Fig.~\ref{fig:unrolled} (b) for image-to-patient registration with CT data to allow drill trajectory pose estimation not only on the basis of endoscopic but also on OCT data. Further on, beside those 2D-3D registration techniques, the obtained imaging depth inside the OCT data is intended for a 3D-3D registration approach with CT data. This will allow additional comparison including determination of the error of both methods.

Due to OCT imaging artifacts related to the catheter structure, e.g., non-uniform rotational distortion, inaccuracies in the reconstructed drill hole occur. On the one hand, a specific torsionally rigid OCT probe with an outer diameter fitted to the drill hole should be used to reduce the artifacts and to simpilfy the OCT image pre-processing. On the other hand, the OCT imaging catheters that were used in this paper are already clinically approved, achieving immediate implementation for MICI cadaver and human studies. 

\paragraph{Conflict of Interest} The authors declare that they have no conflict of interest.

\paragraph{Funding} This research has received funding from the European Union as being part of the ERFE OPhonLas project as well as from the Deutsche Forschungsgemeinschaft (DFG) under grant number KA 2975/4-2 and SCHL 1844/2-2. 

\bibliographystyle{spmpsci} % mathematics and physical sciences
\bibliography{literature}   % name your BibTeX data base

\end{document}